\documentclass[conference]{smartaugment}

\usepackage{cancel}
\usepackage{float}
\usepackage{caption}
\usepackage{subcaption}
\usepackage{xcolor}
\definecolor{red}{HTML}{F9001A}
\definecolor{orange}{HTML}{F96200}

\usepackage{cite}

\ifCLASSINFOpdf
   \usepackage[pdftex]{graphicx}
\else
   \usepackage[dvips]{graphicx}
\fi

\usepackage{amssymb}
\usepackage{commath}

\usepackage{amsmath}
\usepackage{mathtools}

\usepackage{algorithmic}
\usepackage[linesnumbered,ruled,vlined]{algorithm2e}

\newlength\mylen

\usepackage{array}
\usepackage{booktabs}

\usepackage{stfloats}
\usepackage{url}

\makeatletter
\newcommand{\removelatexerror}{\let\@latex@error\@gobble}
\makeatother

\begin{document}
\urlstyle{tt}

\title{Smart(Sampling)Augment: Optimal and Efficient Data Augmentation for Semantic Segmentation}

\author{\IEEEauthorblockN{Misgana Negassi\IEEEauthorrefmark{1}\IEEEauthorrefmark{2}\IEEEauthorrefmark{3}, Diane Wagner\IEEEauthorrefmark{1}\IEEEauthorrefmark{2}, Prof. Dr. Alexander Reiterer\IEEEauthorrefmark{2}\IEEEauthorrefmark{3}}\\

\IEEEauthorblockA{\IEEEauthorrefmark{2} Institute for Sustainable Systems Engineering (INATECH), 
University of Freiburg}
\IEEEauthorblockA{\IEEEauthorrefmark{3} Fraunhofer Institute for Physical Measurement Techniques IPM Freiburg, Germany}
\IEEEauthorblockA{\IEEEauthorrefmark{1} equal contribution}

\thanks{Corresponding author: M.Negassi(email:misgana.negassi@ipm.fraunhofer.de)}}
\thanks{D.Wagner(email:wagnerd@cs.uni-freiburg.de)}
\thanks{A.Reiterer(email:alexander.reiterer@ipm.fraunhofer.de)}

\maketitle
\thispagestyle{plain}
\pagestyle{plain}
\begin{abstract}
Data augmentation methods enrich datasets with augmented data to improve the performance of neural networks. Recently, automated data augmentation methods have emerged, which automatically design augmentation strategies. Existing work focuses on image classification and object detection, whereas we provide the first study on semantic image segmentation and introduce two new approaches: \mbox{\textit{SmartAugment}} and \mbox{\textit{SmartSamplingAugment}}. SmartAugment uses Bayesian Optimization to search over a rich space of augmentation strategies and achieves a new state-of-the-art performance in all semantic segmentation tasks we consider. \mbox{SmartSamplingAugment}, a simple parameter-free approach with a fixed augmentation strategy competes in performance with the existing resource-intensive approaches and outperforms cheap state-of-the-art data augmentation methods. Further, we analyze the impact, interaction, and importance of data augmentation hyperparameters and perform ablation studies, which confirm our design choices behind \mbox{SmartAugment} and \mbox{SmartSamplingAugment}. Lastly, we will provide our source code for reproducibility and to facilitate further research.
\end{abstract}

\IEEEpeerreviewmaketitle

\section{Introduction}

In many real-world applications, only a limited amount of annotated data is available, which is particularly \mbox{pronounced} in medical imaging applications, where expert knowledge is indispensable to annotate data accurately \cite{Negassi.2020},\cite{QuentinPentek.2018}. 
Given \mbox{insufficient} training data, deep learning methods frequently overfit and fail to learn a discriminative function that \mbox{generalizes} well to unseen examples \cite{Goodfellow}. \textit{Data augmentation} is an established approach that improves the generalization of neural networks by adjusting the limited available data to get more and diverse samples for the network to train on.
In most cases, additional data is constructed by simply applying label-preserving transformations to the original data. In image processing, for instance, these can be simple geometric transformations (e.g., rotation), color transformations (e.g., contrast adjustments), or more complex approaches such as CutMix \cite{cutmix}. Data augmentation has been applied to various areas, such as image classification\cite{mixup}, object detection \cite{Detectron2018}, and semi-supervised learning \cite{simclr}. This study provides a first and extensive study on automated data augmentation for semantic segmentation on different and diverse datasets.

Data augmentations used in practice are mostly simple and easy to implement. Despite this simplicity, the choice of augmentations is crucial and requires domain knowledge. \mbox{Recently}, automated data augmentation methods were proposed that learn optimal augmentation policies from data without the need for domain knowledge \cite{autoaugment}, \cite{objdetaugment}, \cite{randaugment}, \cite{population}. These approaches improve performance and show the shortcomings of manually designed data augmentation strategies commonly used across different domains and datasets.

The main focus of current research in automated data augmentation is image classification \cite{randaugment},\cite{autoaugment}, with a particular blind spot being dense prediction tasks like semantic segmentation. Furthermore, these methods either use complicated proxy tasks to learn an optimal augmentation strategy \cite{autoaugment} or optimize the augmentation operations without taking the type of augmentation applied and the probability of their application into account \cite{randaugment}.

\begin{table}[t!]
    \centering
    \begin{tabular}{lccccc}
    \toprule
    Dataset &  Default & Rand++ & Trivial & Smart  & SmartSampling \\
    \midrule
    KITTI & 65.07 &  67.19  & 64.82 & \textbf{68.84}   & 66.53  \\
    Ravenna    &  88.37  & 90.71 &  90.53  & \textbf{91.00} & 90.72 \\
    EM       & 77.25   &  78.83    & 78.15  &  \textbf{79.04}  & 78.42 \\
    Erfasst    & 67.01  & 68.75  & 66.79  & \textbf{73.72}  &  70.24\\
    \midrule
    \# Iterations & 1 & 50 & 1 & 50 & 1 \\
    \bottomrule
    \end{tabular}
    \caption{Test mean Intersection over Union (mIoU) in percentage for different algorithms on semantic segmentation datasets. \mbox{SmartAugment} outperforms all the other data augmentation strategies across all datasets. SmartSamplingAugment competes with the previous state-of-the-art approaches and outperforms TrivialAugment, a comparably cheap method. For \mbox{DefaultAugment}, TrivialAugment, and SmartSamplingAugment, we evaluated each experiment three times using different seeds to get the mean performance. For RandAugment++ and SmartAugment, we took the mean test IoU over the three best-performing validation configurations. Note that RandAugment in this comparison uses Random Search; DefaultAugment represents the baseline, and the higher the value, the better the performance.}
    \label{table:results}
\end{table}

In this work, we introduce two novel data \mbox{augmentation} methods, \textit{SmartAugment} and \mbox{\textit{SmartSamplingAugment}} for semantic segmentation tasks and study them over multiple and diverse datasets from diverse applications: medical imaging, bridge inspection, and autonomous driving. SmartAugment uses Bayesian Optimization \cite{BO},\cite{bohb} to optimize data augmentation strategies and outperforms the previous state-of-the-art methods \mbox{(see Table \ref{table:results})} across all semantic segmentation tasks we consider. In contrast to existing approaches, we define a separate set of each color and geometric data augmentation operations, search for their optimal number of operations and magnitudes, and further optimize a probability $P$ of applying these augmentations.

While SmartAugment performs exceedingly well compared to existing approaches, hyperparameter optimization requires multiple iterations to find the best augmentation strategy, which can be expensive for researchers with computational constraints.
With this in mind, we developed a fast and efficient data augmentation method, SmartSamplingAugment, that has a competitive performance to current best methods and outperforms TrivialAugment \cite{trivialaugment}, a previous state-of-the-art simple augmentation method. SmartSamplingAugment is a parameter-free approach that samples augmentation operations according to their weights, and the probability of application is annealed during training.

We summarize our contributions in the following points:
\begin{itemize}

    \item We provide a first and extensive study of data augmentation on different and diverse datasets for semantic segmentation. 
    
    \item We introduce a new state-of-the-art automated data augmentation algorithm for semantic segmentation based on optimizing the number of applied geometric and color augmentations and their magnitude separately.  Furthermore, we optimize the probability of augmentation, which is crucial according to a hyperparameter importance analysis.

    \item We present a novel parameter-free data augmentation approach that is competitive with the previous state-of-the-art data augmentation strategies and outperforms TrivialAugment, the previous state-of-the-art in this cheap setting, by weighting the applied data augmentation operations and annealing their probability of application.
   
\end{itemize}

To reproduce our results, we will provide our codebase available at \url{https://github.com/MVG-INATECH/SmartAugment}

\section{Related Work}

Data augmentation has been shown to have a considerable impact, particularly on computer vision tasks. Simple augmentation methods such as random cropping, horizontal flipping, random scaling, rotation, and translation have been effective and popular for image classification datasets \cite{alexnet},\cite{resnet}, \cite{pyramid}, \cite{dropconnect}. Other approaches add noise or erase part of an image \cite{cutout},\cite{erase} or apply a convex combination of pairs of images and their labels \cite{mixup}. Other approaches use generative adversarial networks to generate new training data \cite{gan}, \cite{ganaug1}.

Automated augmentation methods focus on learning an optimal data augmentation strategy from data \cite{randaugment}, \cite{autoaugment}. Many recent methods define a set of geometric and color augmentations and their magnitude, where the best augmentation strategy is automatically selected. AutoAugment \cite{autoaugment} uses a search algorithm based on reinforcement learning to find the best data augmentation policy with a validation accuracy as the reward. The search space consists of policies which in turn, have many sub-policies. Each sub-policy contains two augmentation operations, their magnitude, and a probability of application. A sub-policy is selected uniformly at random and applied to an image from a mini-batch. This process has high computational demands, and therefore, it is applied on a proxy task with a smaller dataset and model. The best-found augmentation policy is then applied to the target task. 

Population-Based Augmentation (PBA) \cite{pba} uses a population-based training algorithm \cite{population} to learn a schedule of augmentation policies at every epoch during training. The policies are parameterized to consist of the magnitude and probability values for each augmentation operation. PBA randomly initializes and trains a model with these different policies in parallel. The weights of the better-performing models are cloned and perturbed with noise to make an exploration and exploitation trade-off. The schedule is learned with a child model and applied to a larger model on the same dataset.

Fast AutoAugment \cite{fastautoaug} speeds up the search for the best augmentation strategy with density matching. This method directly learns augmentation policies on inference time augmentations and tries to maximize the match of the distribution between augmented and non-augmented data without the need for child models. The idea is that if a network trained on real data generalizes well on augmented validation data, then the policy that produces these augmented data will be optimal. In other words, the policy preserves the label of the images, thus the distribution of the real data. 

Adversarial AutoAugment \cite{adveraug} optimizes a target network and augmentation policy network jointly on target task in an adversarial fashion. The augmentation policy network generates data augmentations policies that produce hard examples, thereby increasing the target network's training loss. The hard examples force the target network to learn more robust features that improve its generalization and overall performance.

RandAugment \cite{randaugment} uses a much reduced search space than AutoAugment and optimizes two hyperparameters: the number of applied augmentations and the magnitude. RandAugment tunes these parameters with a simple Grid Search \cite{gridsearch} on the target task, therefore, removes the need for a proxy task as is the case in AutoAugment \cite{autoaugment}. The authors argue that this simplification helped the strong performance and efficiency of their approach. 

TrivialAugment \cite{trivialaugment} samples one augmentation and its magnitude uniformly at random from a given set of augmentations and applies it to a given image. This method is efficient, parameter-free, and competes with RandAugment \cite{randaugment} in performance for image classification.

In this work, we introduce two novel (automated) data augmentation methods for semantic segmentation: \mbox{SmartAugment} and \mbox{SmartSamplingAugment}. With hyperparameter optimization, SmartAugment finds optimal data augmentation strategy and SmartSamplingAugment's efficient and parameter-free approach competes with the previous state-of-the-art methods.

\section{Methods}
In this section, we present our data augmentation algorithms: SmartAugment and SmartSamplingAugment. Similar to previous methods, namely RandAugment and \mbox{TrivialAugment}, we define a set of color and geometric augmentations along with their magnitudes as shown in Table \ref{table:augmentations}. We describe our algorithms in detail in the following subsections.

\begin{table}[H]
    \centering
    \begin{tabular}{lclc}
    \toprule
    Color Ops & Range & Geometric Ops & Range \\
    \midrule
     Sharpness & (0.1, 1.9)   &  Rotate & (0, 30) \\ 
     AutoContrast& (0, 1) & ShearX& (0.0, 0.3)\\ 
     Equalize& (0, 1)& ShearY & (0.0, 0.3)\\
     Solarize & (0, 256)  & TranslateX &(0.0, 0.33)  \\
     Color &  (0.1, 1.9)    & TranslateY& (0.0, 0.33)\\
     Contrast & (0.1, 1.9)  &  Identity*  &  \\
     Brightness & (0.1, 1.9)  &  & \\
    \bottomrule
    \end{tabular}
    \caption{Detailed overview of data augmentation operations and their magnitude ranges. We took the same augmentations used in RandAugment\cite{randaugment}. *The Identity operation only belongs to this list for the RandAugment and TrivialAugment approaches.}
    \vspace{-5pt}
    \label{table:augmentations}
\end{table}
\vspace{-10pt}
\subsection{SmartAugment}
\begin{figure}[H]
     \centering
     \begin{subfigure}[b]{0.5\textwidth}
         \centering
         \includegraphics[width=\textwidth]{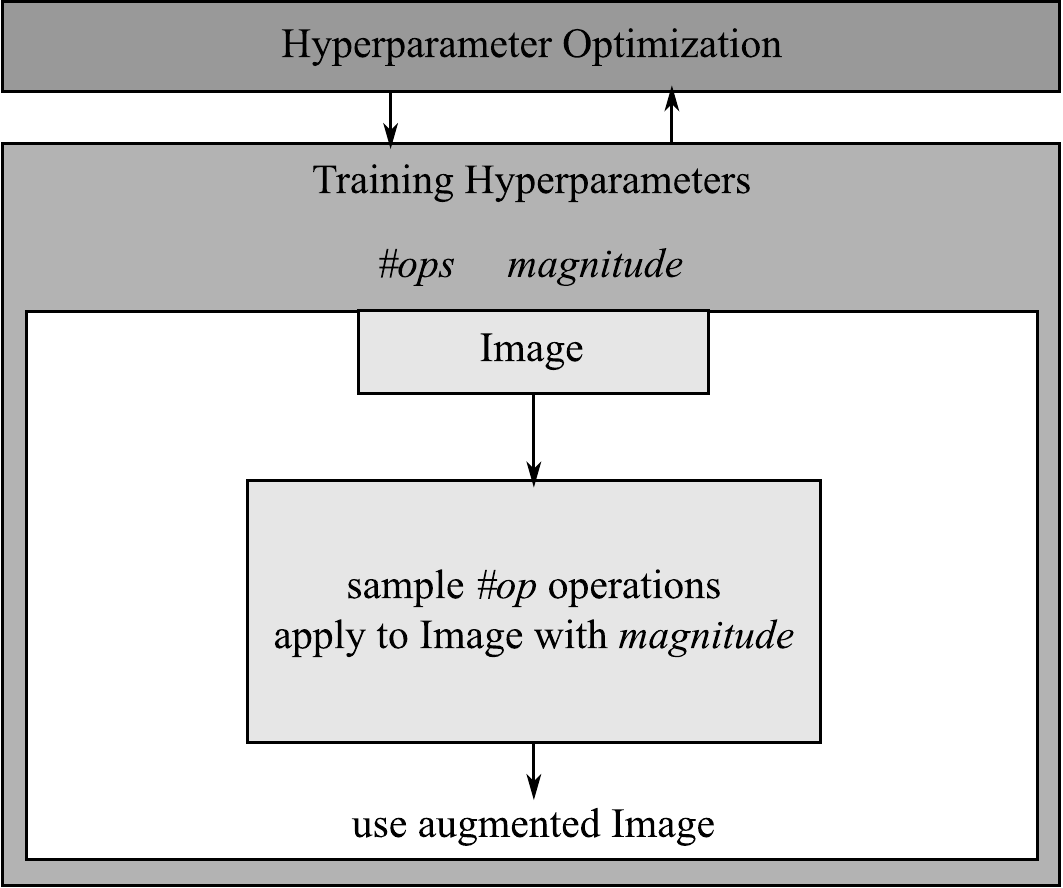}
         \caption{RandAugment}
         \vspace{10pt}
         \label{fig:randops}
     \end{subfigure}
     \hfill
     \begin{subfigure}[b]{0.5\textwidth}
         \centering
         \includegraphics[width=\textwidth]{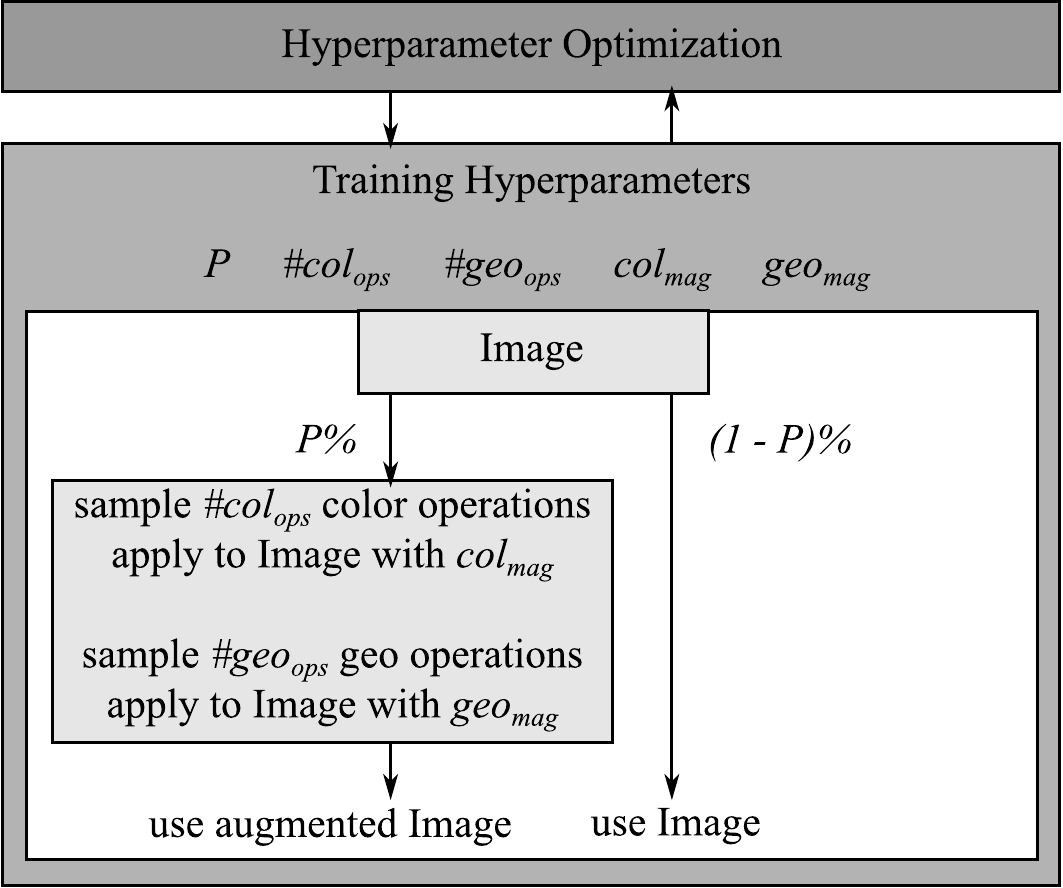}
         \caption{SmartAugment}
         \label{fig:smartops}
     \end{subfigure}
     \hfill
        \caption{Comparison of RandAugment and SmartAugment}
\end{figure}

SmartAugment optimizes the number of sampled color and geometric augmentations and their magnitude separately \mbox{(see Figure \ref{fig:smartops} and Algorithm \ref{alg:smartaugmentalgo})}. Having these distinct sets of augmentations allows control over the type of applied augmentation instead of optimizing the total number of sampled augmentations and their magnitude collectively. SmartAugment also optimizes a parameter that determines the probability of applying data augmentations $P$ instead of having the Identity operation in the augmentation list, as done by recent approaches.

SmartAugment uses Bayesian Optimization (BO) \cite{bohb} to search for the best augmentation strategy. The space of augmentation strategies in SmartAugment has the following parameters: the number of color augmentations $N_{C}$, the number of geometric augmentations $N_{G}$, the color magnitude $M_{C}$, the geometric magnitude $M_{G}$, and the probability of applying augmentations ${P}$. These hyperparameters are fed into the BO algorithm and are optimized until a given budget is exhausted. Once BO chooses the augmentation parameters, the augmentations are sampled from the given list of augmentation operations as listed in Table \ref{table:augmentations}. Here we note that in SmartAugment, each augmentation can be sampled only once per image, while RandAugment allows sampling the same augmentation several times per image.
 
\subsection{SmartSamplingAugment}
\begin{figure}[ht!]
     \centering
     \vspace{34pt}
     \begin{subfigure}[b]{0.5\textwidth}
         \centering
         \includegraphics[width=\textwidth]{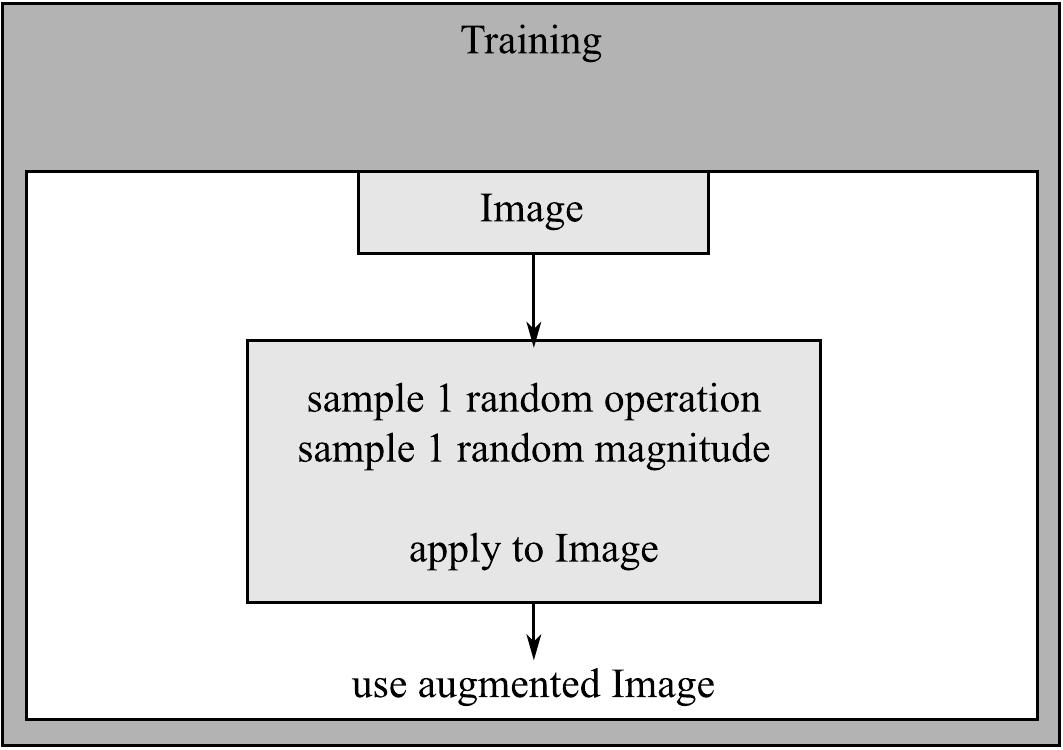}
         \caption{TrivialAugment}
         \vspace{10pt}
         \label{fig:trivialops}
     \end{subfigure}
     \begin{subfigure}[b]{0.5\textwidth}
     \vspace{34pt}
         \centering
         \includegraphics[width=\textwidth]{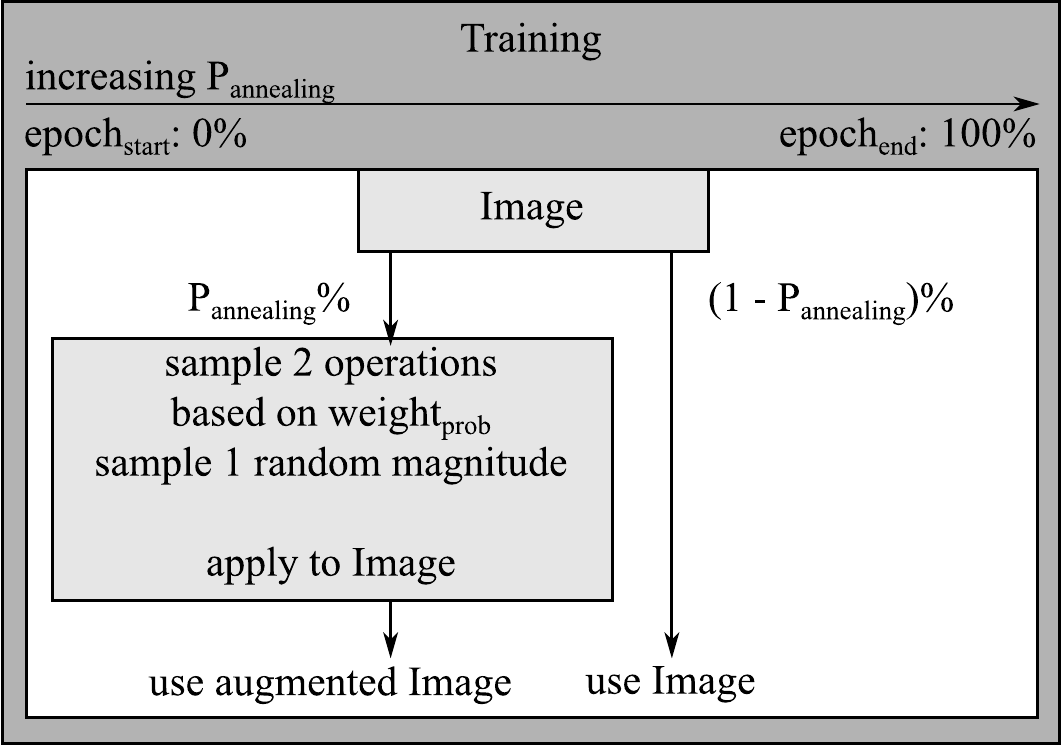}
         \caption{SmartSamplingAugment}
         \label{fig:smartsamplingops}
     \end{subfigure}
     \hfill
        \caption{Comparison of TrivialAugment and SmartSamplingAugment}
\end{figure}

In this section, we present SmartSamplingAugment, a tuning-free and computationally efficient algorithm. The number of sampled augmentations in SmartSampling is fixed to two augmentation operations, and the magnitude is sampled randomly from the interval [5, 30] (see Figure \ref{fig:smartsamplingops} and Algorithm \ref{alg:smartsamplingalgo}). These design choices are based on our preliminary experiments and seem to generalize well to unseen datasets.
SmartSamplingAugment samples augmentations with a probability derived from weights, which we set based on an ablation study for image classification on CIFAR10 from RandAugment \cite{randaugment}. In this study\cite{randaugment}, the average improvement in performance is computed when a particular augmentation operation is added to a random subset of augmentations. We selected the augmentations with a positive average improvement and transformed this value into probabilities, by which we define the weights.

\begin{figure}[b!]
\removelatexerror
\vspace{-20pt}
\begin{algorithm}[H]
\caption{Pseudocode for SmartAugment}
\label{alg:smartaugmentalgo}

\KwIn{Data $D$,
\newline List of color augmentations $C_{LIST}$,
\newline List of geometric augmentations $G_{LIST}$
}
\For{each configuration}
{   Select 5 hyperparameters via BO:\\
    \qquad 1) \# Color augmentations $N_{C}$ to sample,\\
    \qquad 2) \# Geometric augmentations $N_{G}$ to sample, \\
    \qquad 3) Color magnitude $M_{C}$, \\
    \qquad 4) Geometric magnitude $M_{G}$, \\
    \qquad 5) Probability $P$ of applying augmentations \\
    \For{each epoch}
    {
        \For{each image $I$ in D}
        {
            Sample $var$ uniformly from [0, 1]\\
            \If{$var > P$}{
                use $I$  \tcp*{do not augment}
            }
            \Else{
                $C$ := sample $N_{C}$ ops from $C_{LIST}$ \\
                $G$ := sample $N_{G}$ ops from $G_{LIST}$ \\
                $I_{AUG}$ := apply $C$ with $M_{C}$ \\
                \qquad \qquad and $G$ with $M_{G}$ on $I$ \\
                use $I_{AUG}$
            }
        }
    }
}
\end{algorithm}

\begin{algorithm}[H]
\caption{Pseudocode for SmartSamplingAugment}
\label{alg:smartsamplingalgo}

\KwIn{Data $D$,
\newline List of augmentations $A:=[a_1, a_2, \dots, a_{-1}]$,
\newline Weights $W:=[w_{a_1}, w_{a_2}, \dots , w_{a_{-1}}]$
}
\For{each epoch}
{
    Update $P$ \tcp*{$P$ is linearly annealed}
    \For{each image $I$ in $D$}
    {
        Sample $var$ from [0, 1]\\
        \If{$var > P$}{
            use $I$  \tcp*{do not augment}
        }
        \Else{
            $A_{W}$ := sample 2 ops from $A$ based on $W$ \\
            $M$ := sample magnitude from [5, 30]\\
            $I_{AUG}$ := apply $A_{W}$ with $M$ on $I$ \\
            use $I_{AUG}$
        }
    }
}
\end{algorithm}

\end{figure}

In SmartSampling, we linearly anneal the parameter $P$, that determines the probability of applying data augmentations, from \mbox{0 to 1}, increasing the percentage of applying augmentation over the whole training epochs. That way, the model first sees the original data in the early epochs and encounters more variations as the training progresses.


\section{Experiments and Results}

In this section, we empirically evaluate and analyze the performance of SmartAugment and SmartSamplingAugment on several datasets and compare it to the previous state-of-the-art approaches.
Furthermore, we investigate the impact, interaction, and importance of the optimized data augmentation hyperparameters. 

\subsection{Experimental Setup}
\paragraph*{\textbf{RandAugment++}}
Classical RandAugment \cite{randaugment} uses simple Grid Search \cite{gridsearch} to optimize its hyperparameters. Evaluating the full grid of classical RandAugment would lead to evaluate nearly 100 iterations (31 $\times$ 3 iterations: magnitude in the range of [0, 30] and the number of operations in the range of \mbox{[1, 3]}) which is computationally very expensive. Therefore, we decided to implement an updated version of RandAugment, which we call \mbox{\textit{RandAugment++} (see Figure \ref{fig:randops})} that uses the same algorithm but optimizes its hyperparameters with Random Search. \mbox{Random Search} is known to perform better than Grid Search \cite{randomsearch} and its number of iterations is not limited to the size of the grid in the search space. Furthermore, using Random Search enabled us to reduce the number of iterations and increase the search space for RandAugment++ without additional computational costs. We analyzed the performance of RandAugment++ with different operations on the EM dataset and found out that constraining the number of applied operations to 3 is not optimal (see Figure \ref{fig:randops-em}). From this observation, we increase the upper limit for the number of applied operations from 3 to 16, which denotes the total number of augmentations in the list we sample from. In order to ensure comparability between RandAugment++ and SmartAugment, we use the same computational budget of \mbox{50 iterations} for both methods. We show in an ablation study that RandAugment++ is a better choice than classical RandAugment.

\paragraph*{\textbf{Default and TrivialAugment}}
For completeness we \mbox{include} in our experiment a standard augmentation strategy, we named \textit{DefaultAugment} and use it as our baseline. This default augmentation strategy is commonly used in semantic segmentation literature \cite{hrnet},\cite{deeplab} and uses the following standard augmentations: horizontal flipping $(p_{flip}$=$0.5)$, random rotation $(range$=$[-45, 45])$, random scaling $(range$=$[-0.35, 0.35])$, where $p_{flip}$ represents the probability of applying this particular augmentation. Further, we re-implemented another recent method \mbox{TrivialAugment} (see Figure \ref{fig:trivialops}), for semantic segmentation  and included it in our experiments. 

\paragraph*{\textbf{Training Setup}} For all experiments, we use the U-Net architecture \cite{unet} to train our models and split our datasets into training, validation, and test set. To find good fitting training hyperparameters (e.g., learning rate and weight decay) for our in-house datasets, we performed Random Search over ten configurations until model convergence. We apply random crop or downsize with a 50\% probability before passing the data to the different augmentation strategies for efficient memory and compute use. For KITTI and EM datasets, we use a similar training setup as in \cite{deeplab}. To reduce memory requirements, we use mixed precision training with 16 bits. For our experiments, we made use of four GeForce GTX 1080 GPUs. For better reproducibility, we list the training parameters for each of the datasets for a detailed view in Table~\ref{table:train_params}.  

Furthermore, we performed early stopping on the validation set. In order to save compute and still get enough samples on the validation set for early stopping, we evaluate every 10\% of the epochs on the validation set. This ensures that for each dataset, independent of the number of epochs needed until convergence, the number of epochs evaluated on the validation set is proportional to the total number of epochs. We run these experiments three times for the different data augmentation approaches and take the mean of the test IoU to ensure a fair comparison. In the case of RandAugment and SmartAugment, we evaluated 50 configurations for each method and report the mean test IoU of the three best performing configurations.

For all our experiments, we use Stochastic Gradient Descent (SGD) optimizer \cite{SGD} and Cosine Annealing \cite{sgdr} as our learning rate scheduler, and anneal the learning rate over the total number of epochs. 

\begin{figure}[H]
\centering
\vspace{-5pt}
\begin{minipage}[b]{.5\textwidth}
    \begin{subfigure}[b]{0.225\textwidth}
         \centering
         \includegraphics[width=\textwidth]{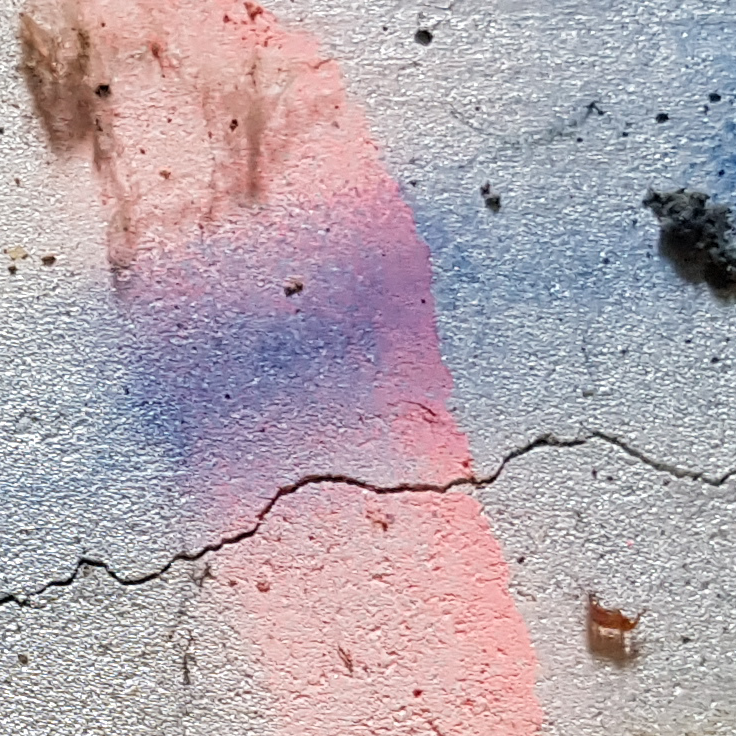}
         \subcaption{Erfasst}
         \label{fig:erfasst}
    \end{subfigure}
     \hfill
     \begin{subfigure}[b]{0.75\textwidth}
         \centering
         \includegraphics[width=\textwidth]{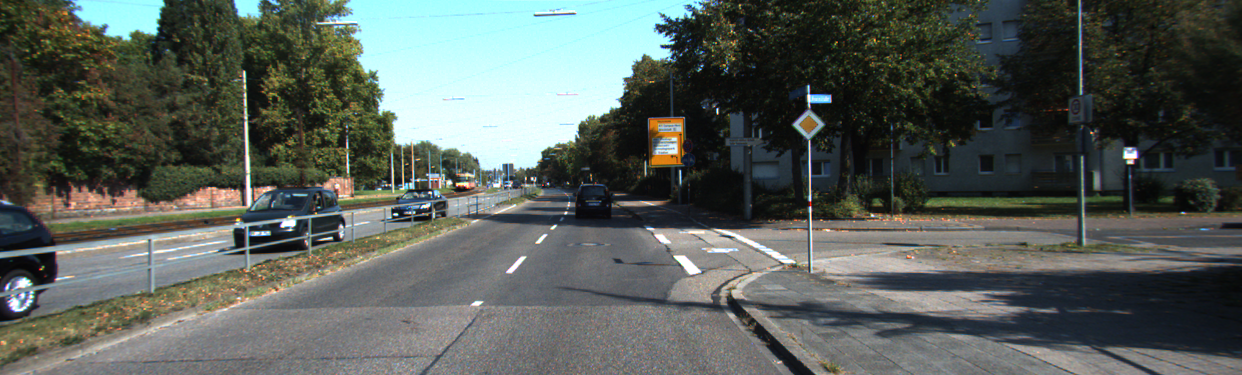}
         \subcaption{KITTI}
         \label{fig:kitti}
     \end{subfigure}
     \hfill
\caption{Infrastructure mapping datasets}\label{label-a}
\end{minipage}\qquad

\begin{minipage}[b]{.5\textwidth}
\vspace{15pt}
    \begin{subfigure}[b]{0.62\textwidth}
         \centering
         \includegraphics[width=\textwidth]{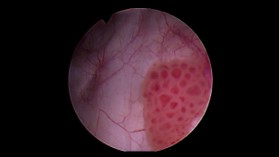}
         \caption{Ravenna}
         \vspace{-10pt}
         \label{fig:ravenna}
     \end{subfigure}
    \begin{subfigure}[b]{0.35\textwidth}
         \centering
         \includegraphics[width=\textwidth]{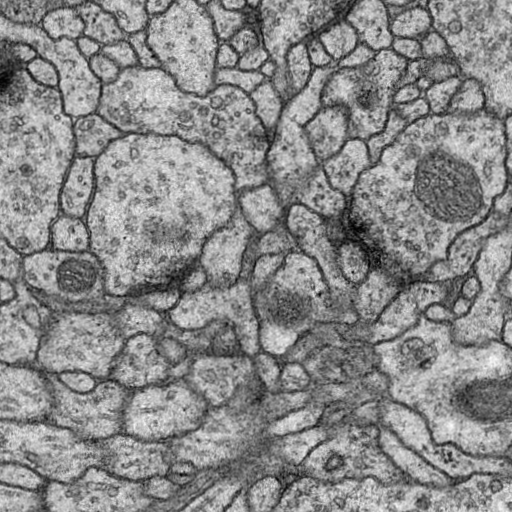}
         \caption{EM}
         \vspace{-10pt}
         \label{fig:em}
     \end{subfigure}
     \hfill
     
     \hfill
\caption{Biomedical datasets}
\end{minipage}
\end{figure}

\paragraph*{\textbf{Datasets}}
We evaluate all approaches on datasets from diverse applications: medical imaging (Ravenna \cite{Negassi.2020}: Fig. \ref{fig:ravenna}, the 2D EM segmentation challenge dataset (EM) \cite{EM}: \ref{fig:em}), bridge inspection \mbox{(Erfasst: Fig. \ref{fig:erfasst})} and autonomous driving (KITTI \cite{kitti}: Fig. \ref{fig:kitti}) to achieve meaningful results. These datasets differ in size, resolutions, and type of images (RGB natural images, Grayscale). Since Ravenna and Erfasst are highly class-imbalanced datasets, we use a weighted cross-entropy loss during training. The weights are computed beforehand with inverse frequency of the number of pixels belonging to a specific class in the training set.

\begin{table}[H]
\centering
\begin{tabular}{ lccccc}
\toprule
Dataset & \# Data  &   Resolution  & Batch size & Learning rate & Epochs \\
 \midrule
KITTI & 200 & 185x612 &  4 & 0.001  &4000 \\
EM & 30 &512x512 & 2& 0.01 & 500\\
Ravenna & 1684 &180x180 & 3& 0.001 & 2000\\
Erfasst & 50 & 864x864 & 2& 0.05 & 5000\\

\bottomrule
\end{tabular}
\caption{Training parameters for each dataset.}
\label{table:train_params}
\end{table}

\subsection{Comparison to the state-of-the-art}

In Table \ref{table:results} we compare our methods, SmartAugment and SmartSamplingAugment, to the aforementioned data augmentation methods as well as to our baseline, DefaultAugment, a basic augmentation strategy that is commonly used in semantic segmentation literature \cite{deeplab},\cite{hrnet}. SmartAugment outperforms the previous state-of-the-art methods across all datasets, while SmartSamplingAugment competes with the previous state-of-the-art methods and outperforms the comparably cheap augmentation method, TrivialAugment. Moreover, SmartSamplingAugment outperforms RandAugment++ on half of the datasets, even though the latter has 50 times more budget.
\subsection{Analysis with fANOVA}
In this section, we analyze the impact, interaction, and importance of augmentation hyperparameters across different datasets with fANOVA \cite{fanova}. We quantify and visualize the effect of different augmentation configurations on the overall model performance on the validation mean IoU metric.

\begin{figure}[t!]
     \centering
     \begin{minipage}[b]{.5\textwidth}
     \begin{subfigure}[b]{1.0\textwidth}
         \centering
         \includegraphics[width=\textwidth]{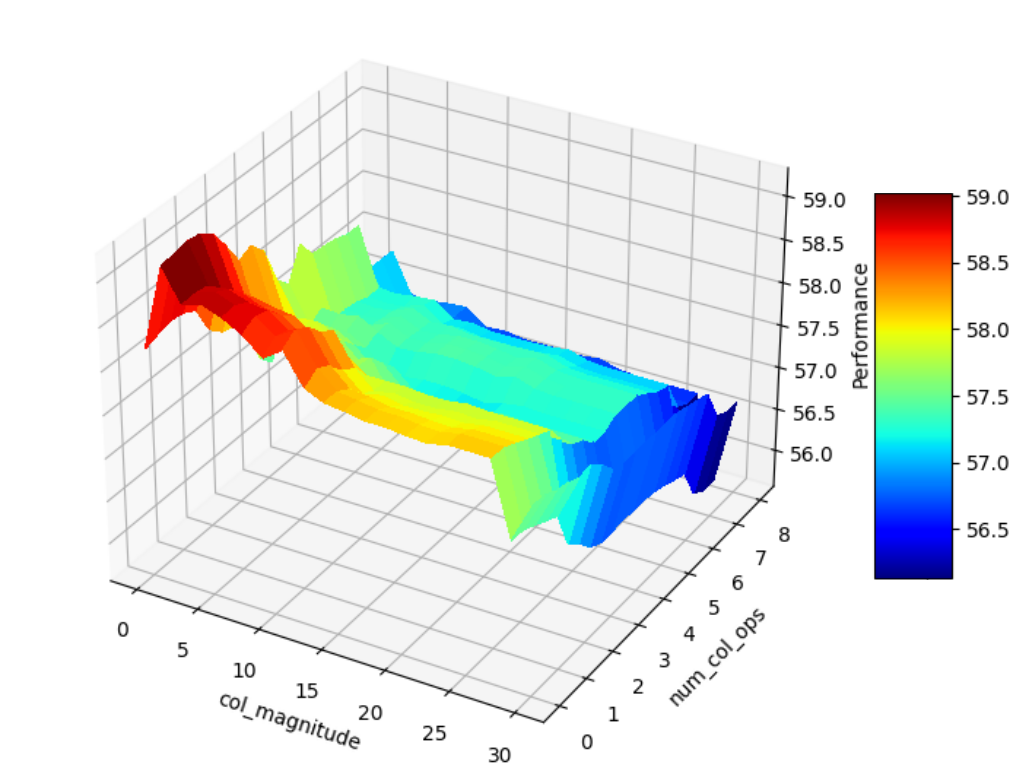}
         \caption{SmartAugment on KITTI}
         \vspace{10pt}
         \label{fig:smartkittimag}
     \end{subfigure}
     \hfill
     \begin{subfigure}[b]{1.0\textwidth}
         \centering
         \includegraphics[width=\textwidth]{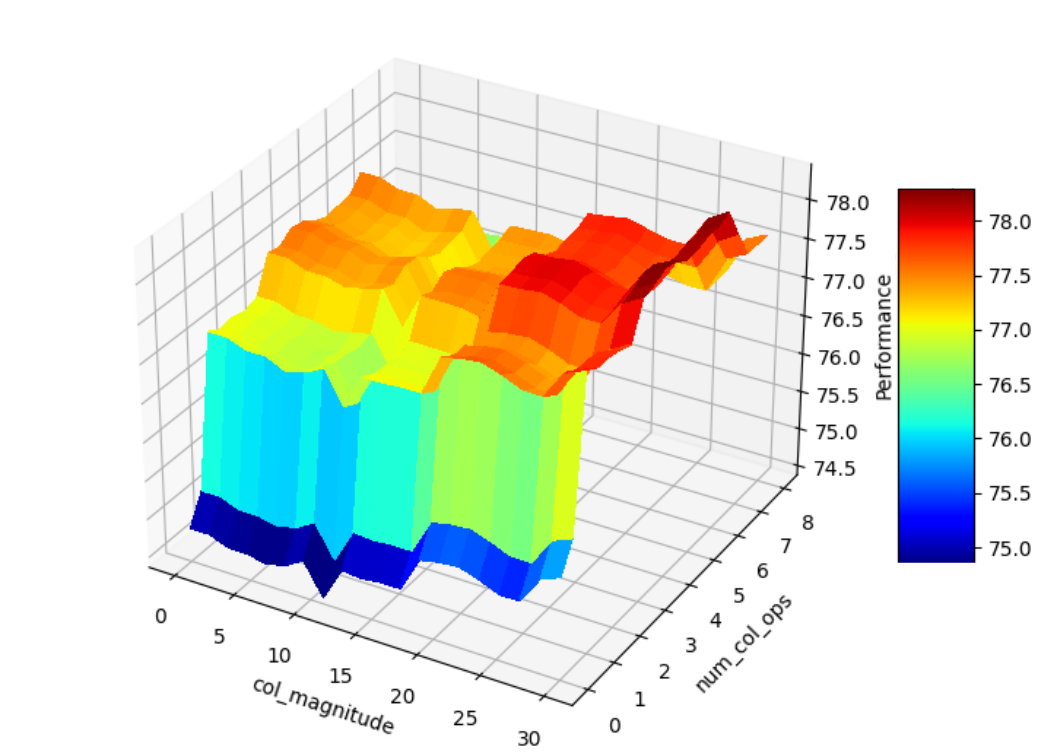}
         \caption{SmartAugment on EM}
         \label{fig:smartemmag}
     \end{subfigure}
     \hfill
    \caption{The impact of hyperparameters on different datasets. This figure shows that the good values for each hyperparameter depend on the dataset.}
    \label{fig:smartkittiemcomp}
    \end{minipage}\qquad
   
\end{figure}

\paragraph*{\textbf{Impact of hyperparameters across different datasets}}
The results in Figure \ref{fig:smartkittiemcomp} show that the optimal strategy of augmentation hyperparameters is dataset-specific and predominantly impacts the overall performance: Applying many color operations with a high color magnitude to the data can be good for the EM dataset but have a detrimental effect on the performance of the KITTI dataset. Therefore, as shown in Figure \ref{fig:smartkittiemcomp}, there are areas in the augmentation space where it is sub-optimal to sample from for a particular dataset but are good for another one.

\begin{figure}[t!]
     \centering
     \begin{minipage}[b]{.5\textwidth}
     \begin{subfigure}[b]{1.0\textwidth}

         \centering
         \includegraphics[width=\textwidth]{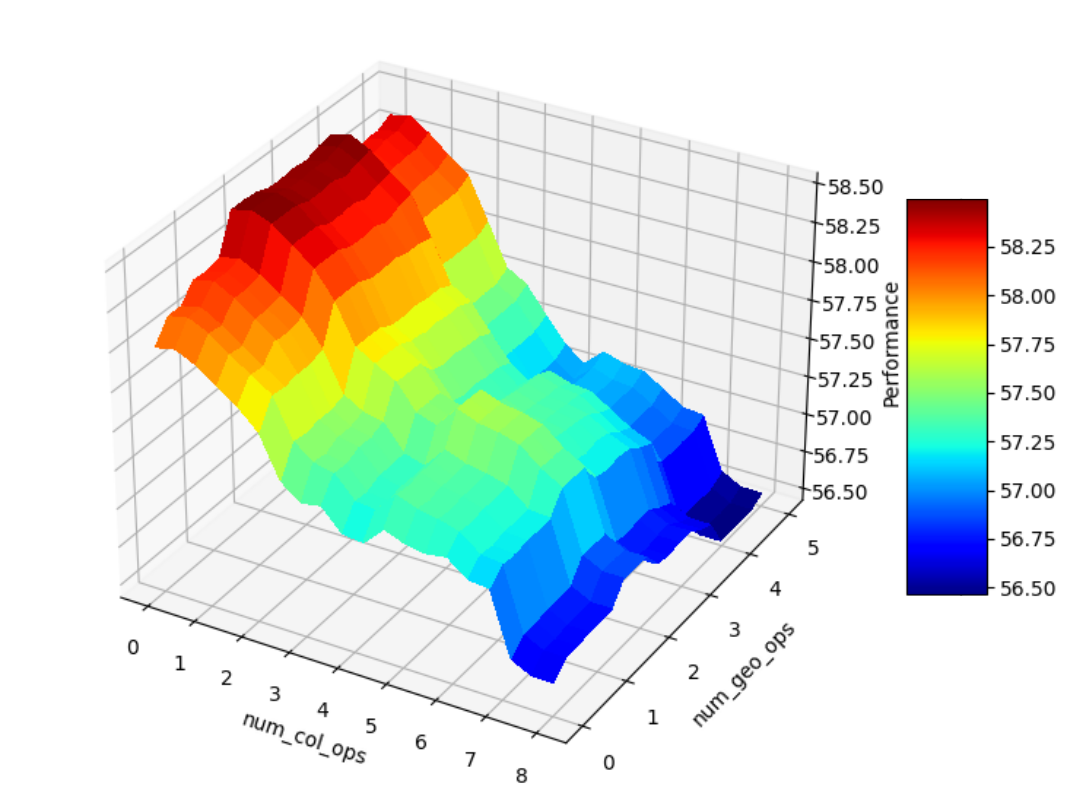}
         \caption{SmartAugment on KITTI}
         \vspace{40pt}
         \label{fig:smartopsfanova}
     \end{subfigure}
     \hfill
     \begin{subfigure}[b]{1.0\textwidth}
         \centering
         \includegraphics[width=0.8\textwidth]{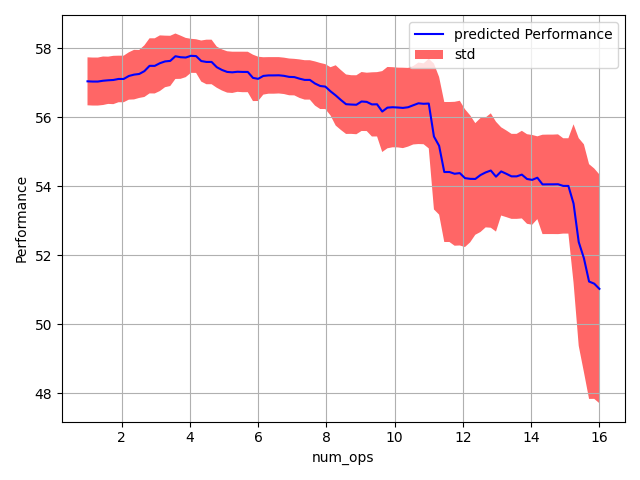}
         \caption{RandAugment++ on KITTI}
         \label{fig:randopsfanova}
     \end{subfigure}
     \hfill
        \caption{Comparison considering the number of applied augmentations: RandAugment++ optimizes the total number of augmentations, whereas SmartAugment differs between the number of color augmentations and geometric augmentations. This figure shows that the performance of a total number of augmentations depends on the types of augmentations.}
        \vspace{-5pt}
    \label{fig:randsmartcompfanova}
    \end{minipage}\qquad
\end{figure}

\paragraph*{\textbf{Hyperparameter Interaction Analysis}}
Furthermore, we analyze the interaction of hyperparameters and their effect on model performance. As mentioned in the Methods section, SmartAugment optimizes the color and geometric augmentations separately. The results in Figure \ref{fig:randsmartcompfanova} confirm our hypothesis that this is a good design choice.  At a closer look, the figure shows that for optimal performance, it does not suffice to optimize the number of applied augmentations; rather, it is crucial to sample the right type of augmentation from the augmentation list carefully. For instance, according to Figure \ref{fig:randopsfanova}, choosing four operations from the total number of augmentation seems to be the optimal choice for the KITTI dataset. However, according to Figure \ref{fig:smartopsfanova}, just sampling \textquotedblleft blindly" four augmentation operations from the entire augmentation list might not always be a good choice. If we would pick four color augmentation operations and zero geometric augmentation operations, the performance would be significantly sub-optimal. 

\paragraph*{\textbf{Hyperparameter Importance Study}}
In many algorithms that have a large hyperparameter space, only a few parameters are usually responsible for most of the performance improvement \cite{fanova}. In this study, we use fANOVA to quantify how much each hyperparameter contributes to the overall variance in performance. As we observe in Table \ref{table:param_importance}, the importance of a specific hyperparameter strongly depends on the dataset. For instance, the geometric magnitude has a much higher impact on the KITTI dataset than other datasets. Moreover, the results from Table \ref{table:param_importance} show that optimizing the probability of application is an important design choice since this parameter is the most important one in half of the datasets studied in these experiments.
\begin{table}[H]
    \centering
    \begin{tabular}{lccccc}
    \toprule
    Dataset     &p(aug) & col\_mag  & geo\_mag & \#col\_ops &  \#geo\_ops \\
    \midrule
    KITTI               &  0.13   &  0.12 & \textbf{0.24}  & 0.14    & 0.03 \\
    Ravenna             &  \textbf{0.46}   &  0.04 & 0.06  & 0.06    & 0.05 \\
    EM                  &  \textbf{0.25}   & 0.14  &  0.09 &  0.04   & 0.03 \\
    Erfasst             &  0.1    & 0.12  & 0.04  & \textbf{0.22}    & 0.04 \\
    \bottomrule
    \end{tabular}
    \caption{Hyperparameter importance study for different hyperparameters across different datasets on SmartAugment experiments. For instance, for the Ravenna dataset, the probability of applying a data augmentation strategy is responsible for 46\% of mean IoU's variability across the configuration space. The higher the importance value, the more potential it has to improve the performance for a given dataset.}
    \label{table:param_importance}
\end{table}

\subsection{Ablation Studies}
In addition to comparing our methods to the state-of-the-art approaches and the baseline, we report some ablation studies that give deeper insights into the impact of our methods.

\paragraph*{\textbf{RandAugment(++) ablation studies}}
To confirm that the improvement of SmartAugment over RandAugment++ comes from the method differences, we study RandAugment with different optimization methods. For this purpose, we compare classical RandAugment with Grid Search, RandAugment++ with Random Search, and RandAugment with Bayesian Optimization as optimization algorithms. We chose a cheap dataset (EM) for this ablation study. As the results in Table \ref{table:ablation} confirm, \mbox{SmartAugment} outperforms RandAugment, independent of the selected hyperparameter optimization algorithm. 
An interesting observation from the study is that RandAugment++ improves over the classical RandAugment as shown in Table \ref{table:ablation}. It is worthy to note that these improvement gains were achieved with fewer iterations and without additional computational costs. 

\begin{table}[H]
    \centering
    \begin{tabular}{llcc}
    \toprule
    Method & HPO algorithm & EM dataset  & \#iterations\\
    \midrule
    Rand   &Grid Search (classic approach)  & 78.54  & 93 \\
    Rand++ & Random Search                  & 78.83  & 50 \\
    Rand++ & Bayesian Optimization          & 78.84  & 50\\  
    Smart & Bayesian Optimization           & \textbf{79.04} & 50 \\
    \bottomrule
    \end{tabular}
    \caption{Comparison of variants of RandAugment with SmartAugment on the EM dataset. For each of the results, we took the mean of the best three performing configurations. The results show that SmartAugment outperforms RandAugment(++), independent of the hyperparameter optimization algorithm.}
    \label{table:ablation}
\end{table}

\begin{figure}[t!]
         \centering
         \includegraphics[width=0.42\textwidth]{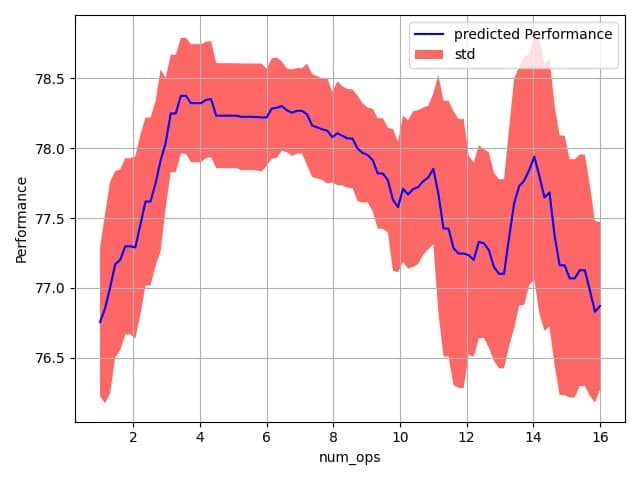}
         \caption{Performance analysis for different numbers of operations for RandAugment++ on EM. These results indicate that the number of applied operations should optimally not be limited to three, as in classical RandAugment.}
         \label{fig:randops-em}
\end{figure}

Further, the results in Figure \ref{fig:randops-em} show that it may be sup-optimal to limit the number of applied augmentation to three, as it is done in classical RandAugment. Therefore, increasing the search space as we do it in RandAugment++ seems to be a good design choice.

\paragraph*{\textbf{SmartSamplingAugment ablation studies}} 

In these ablation studies, we analyze the impact of annealing the probability hyperparameter over the epochs and weighting the augmentation operations. For the experiments without annealing, we set the probability of augmentation $P$ to 1.

The results in Table \ref{table:ablation-smartsampling} show that for three out of four datasets, annealing as well as weighting the augmentations are good design choices. Additionally, Table \ref{table:ablation-smartsampling} shows that the combination of annealing the probability of augmentation and weighting the augmentations for Ravenna and Erfasst datasets improves the performance drastically. Overall, SmartSamplingAugment does comparatively well and outperforms DefaultAugment and TrivialAugment across all datasets (see Table \ref{table:results}). 

\begin{table}[H]
    \centering
    \begin{tabular}{lcc|cc}
    \toprule
    Dataset &  \multicolumn{2}{c|}{weighting}  & \multicolumn{2}{c}{without weighting}\\
    {}   & annealing  & no-annealing&  annealing & no-annealing \\
    \midrule
    KITTI   & 66.53 & 67.15 & \textbf{67.49} & 67.13\\
    Ravenna &\textbf{90.72} & 87.07 & 85.65& 85.68\\
    EM      & 78.52 & \textbf{79.26} & 77.94 & 78.47  \\
    Erfasst & \textbf{70.24} & 68.27 & 64.99 & 64.51\\
    \bottomrule
    \end{tabular}
    \caption{SmartSamplingAugment ablation study analyzing the impact of weighting the augmentations and annealing the probability hyperparameter over the whole epochs for different datasets. We evaluated each experiment three times using different seeds to get the mean. For the experiments without annealing, we set the probability $P$ of applying augmentations to 1.}
    \label{table:ablation-smartsampling}
\end{table}

\begin{figure}[t!]
     \centering
     \begin{subfigure}[b]{.42\textwidth}
         \centering
         \includegraphics[width=\textwidth]{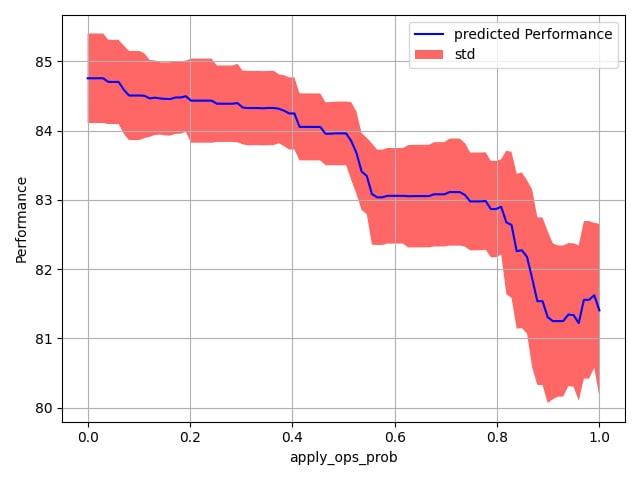}
         \caption{SmartAugment on Ravenna}
         \vspace{10pt}
         \label{fig:smart_ravenna_p}
     \end{subfigure}
     \hfill
     \begin{subfigure}[b]{.42\textwidth}
         \centering
         \includegraphics[width=\textwidth]{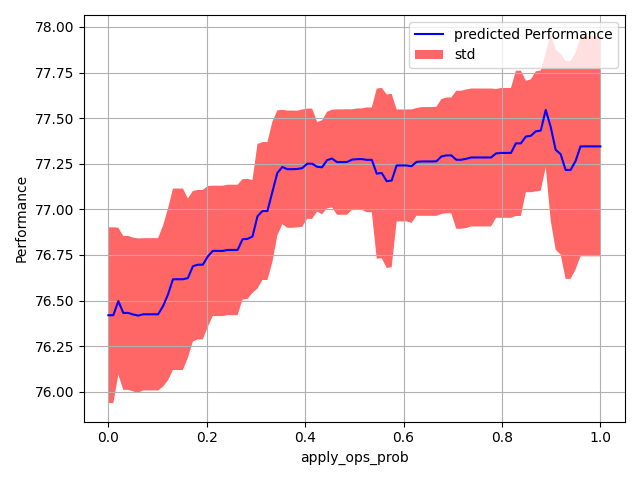}
         \caption{SmartAugment on EM}
         \label{fig:smart_em_p}
     \end{subfigure}
     \hfill
    \caption{Comparison of the probability hyperparameter of applying data augmentations. These results indicate that the EM dataset needs much more data augmentation than the Ravenna dataset.}
    \label{fig:p_comparison}
   
\end{figure}

In the following, we want to give some possible explanations why annealing the augmentations for the EM dataset and weighting the augmentations for the KITTI dataset might not perform well.
According to the hyperparameter importance study in \mbox{Table \ref{table:param_importance}}, the probability of augmenting the data is the most important one for the EM dataset. Figure \ref{fig:smart_em_p}, indicates that the EM dataset benefits from a high percentage of data augmentation; and therefore progressively increasing the probability of applied augmentations can be suboptimal. 

For the Ravenna dataset, the probability hyperparameter of augmenting the data is also the most important parameter according to Table \ref{table:param_importance}. Figure \ref{fig:smart_ravenna_p} shows that a high probability of augmenting data hurts performance for the Ravenna dataset. Therefore, annealing the augmentations for this particular dataset might have a positive impact.

\begin{figure}[H]
         \centering
         \includegraphics[width=0.42\textwidth]{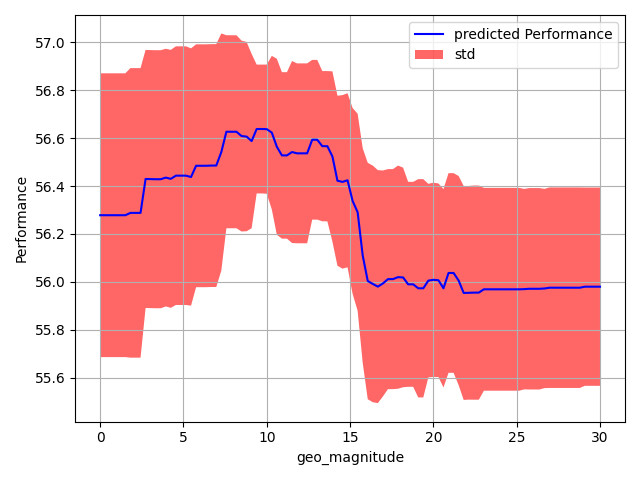}
         \caption{Results from SmartAugment on the KITTI dataset indicate that the geometric magnitude, which is the most important hyperparameter for this particular dataset, should be low. Taking this into account gives a possible explanation for why weighting the data augmentation with weights that focus on geometric augmentations might hurt performance for the KITTI dataset.}
         \label{fig:kitti_geo_mag}
\end{figure}

\begin{figure}[t!]
\centering
\begin{minipage}[b]{.5\textwidth}
    \begin{subfigure}[b]{0.75\textwidth}
         \centering
         \includegraphics[width=\textwidth]{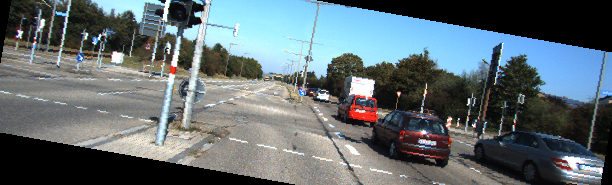}
         \caption{KITTI}
         \vspace{-10pt}
     \end{subfigure}
    \begin{subfigure}[b]{0.225\textwidth}
         \centering
         \includegraphics[width=\textwidth]{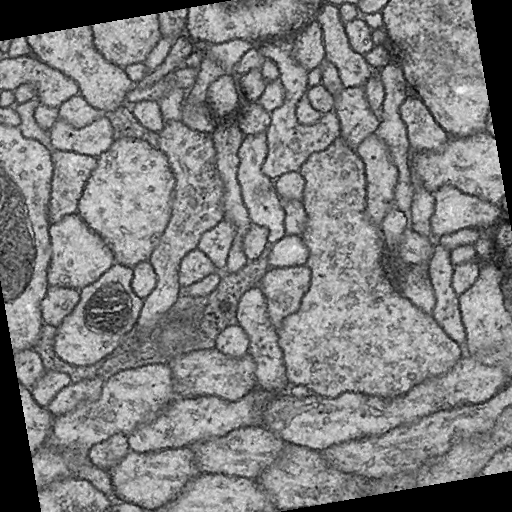}
         \caption{EM}
         \vspace{-10pt}
     \end{subfigure}
     \hfill
     
     \hfill
\caption*{Rotation with magnitude 10}
\end{minipage}

\begin{minipage}[b]{.5\textwidth}
\vspace{10pt}
    \begin{subfigure}[b]{0.75\textwidth}
         \centering
         \includegraphics[width=\textwidth]{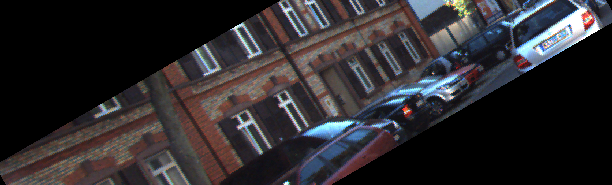}
         \caption{KITTI}
         \vspace{-10pt}
     \end{subfigure}
    \begin{subfigure}[b]{0.225\textwidth}
         \centering
         \includegraphics[width=\textwidth]{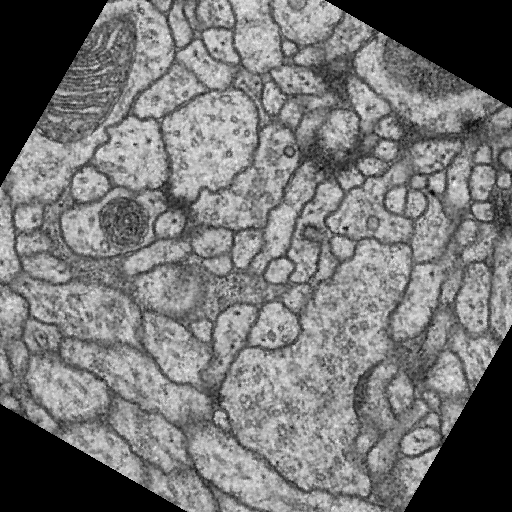}
         \caption{EM}
         \vspace{-10pt}
     \end{subfigure}
     \hfill
     
     \hfill
\caption*{Rotation with magnitude 30}

\vspace{10pt}
    \begin{subfigure}[b]{0.75\textwidth}
         \centering
         \includegraphics[width=\textwidth]{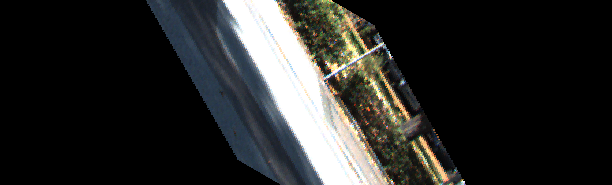}
         \caption{KITTI}
         \vspace{-10pt}
     \end{subfigure}
    \begin{subfigure}[b]{0.225\textwidth}
         \centering
         \includegraphics[width=\textwidth]{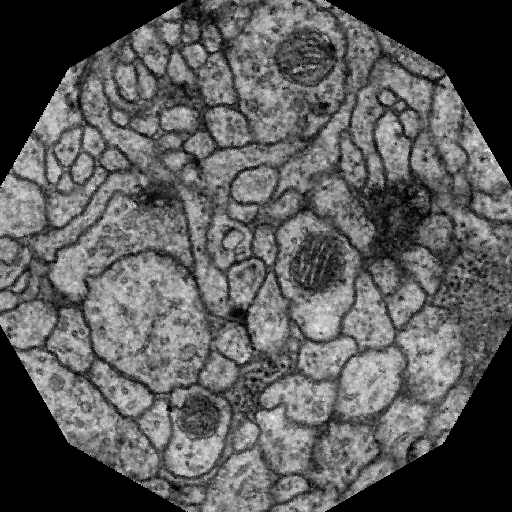}
         \caption{EM}
         \vspace{-10pt}
     \end{subfigure}
     \hfill
     
     \hfill
\caption*{2$\times$ Rotation with magnitude 30}
\end{minipage}
\caption{Visualization of three images from the KITTI and EM datasets, each rotated with a different magnitude.}
\label{fig:rotate}
\end{figure}

Furthermore, Table \ref{table:param_importance} indicates that for the KITTI dataset, the geometric magnitude is the most important hyperparameter, and sampling a high geometric magnitude can hurt performance (see Figure \ref{fig:kitti_geo_mag}). In SmartSamplingAugment, rotation is strongly weighted, and there is a considerable probability that a higher magnitude for this operation is sampled. \mbox{Figure \ref{fig:rotate}} visualizes three KITTI and EM images, each rotated with a different magnitude, and gives an intuition why augmenting the KITTI dataset with high geometric operations can have a detrimental effect on performance.

We note that we select the weights based on a study performed on a classification dataset, which probably is sub-optimal for semantic segmentation. However, this gives insight that studies focusing on optimizing the weights for augmentation operations can be a next step for further research.

\section {Conclusion}
In this work, we provide a first and extensive study of data augmentation for segmentation and introduce two novel approaches: SmartAugment, a new state-of-the-art method that finds the best configuration for data augmentation with hyperparameter optimization, and SmartSamplingAugment, a parameter-free, resource-efficient approach that performs competitively with the previous state-of-the-art approaches. Both methods achieve excellent results on different and diverse datasets.

With SmartAugment, we show that Bayesian Optimization can effectively find an optimal augmentation strategy from a search space where the number of colors and geometric augmentations and their magnitudes are optimized separately, along with a probability hyperparameter for applying data augmentations. Our results show that the type of applied augmentation is essential in making good decisions for improved performance. Furthermore, a hyperparameter importance study indicates that the probability of applying a data augmentation strategy could have considerable responsibility for the mean IoU's variability across the configuration space.

With SmartSamplingAugment, we develop a simple and cheap algorithm that weights the augmentations and anneals augmentations to increase the percentage of augmented images systematically. The results show that this is a powerful and efficient approach that is competitive to the more resource-intensive approaches and outperforms TrivialAugment, a comparably cheap method. Further, SmartSamplingAugment opens the gate for more research on weighting and annealing data augmentation.

In future work, it would be interesting to perform a study that optimizes the weights of data augmentation operation to find an optimal strategy for a segmentation task.

\section*{Acknowledgment}
 The authors would like to thank the Department of Urology of the Faculty of Medicine in University of Freiburg for annotation of cystoscopic images that were used to build the Ravenna (Ravenna 4pi) dataset. Also thanks to Dominik Merkle for providing us with the Erfasst (ErfASst) dataset.

\section*{Funding}
The project is funded by the German Federal Ministry of Education and Research (13GW0203A) and approved by the local Ethical Committee of the University of Freiburg, Germany.
\section*{Conflict of Interest}
The authors declare that they have no conflict of interest.

\bibliography{bibtex/IEEEabrv.bib,bibtex/IEEEexample.bib, bibtex/smartaugment.bib}{}
\bibliographystyle{IEEEtran}

\end{document}